\documentclass{article}
\usepackage{graphicx} 
\usepackage{caption}
\usepackage{subcaption}
\usepackage{url}

\title{Training-free Source Attribution of AI-generated Images via Resynthesis}

\begin{centering}
\author{
Pietro Bongini$^{\star c}$,
Valentina Molinari$^{\star \dagger}$,
Andrea Costanzo$^{\star}$\\
Benedetta Tondi$^{\star}$,
Mauro Barni$^{\star}$\\
$^{\star}$ \textit{University of Siena, Department of Information Engineering and Mathematics}\\
\textit{Siena, Italy}\\
$^{\dagger}$ \textit{IMT School, Lucca, Italy}\\
$^{c}$ \textit{Corresponding author: pietro.bongini@unisi.it}
}
\end{centering}

\date{}

\begin{document}

\maketitle

\begin{abstract}
Synthetic image source attribution is a challenging task, especially in data scarcity conditions requiring few-shot or zero-shot classification capabilities. We present a new training-free one-shot attribution method based on image resynthesis. A prompt describing the image under analysis is generated, then it is used to resynthesize the image with all the candidate sources. The image is attributed to the model which produced the resynthesis closest to the original image in a proper feature space. We also introduce a new dataset for synthetic image attribution consisting of face images from commercial and open-source text-to-image generators. The dataset provides a challenging attribution framework, useful for developing new attribution models and testing their capabilities on different generative architectures. The dataset structure allows to test approaches based on resynthesis and to compare them to few-shot methods. 
Results from state-of-the-art few-shot approaches and other baselines show that the proposed resynthesis method outperforms existing techniques when only a few samples are available for training or fine-tuning. The experiments also demonstrate that the new dataset is a challenging one and represents a valuable benchmark for developing and evaluating future few-shot and zero-shot methods.
\end{abstract}

\section{Introduction}
\label{sec:introduction}
Generative AI has recently known a very fast development. In particular, a rapid evolution of image generative models was driven by several major elements of novelty, including the introduction of diffusion models and the consequent proliferation of text-to-image generators \cite{zhang2023text}. On the one hand, this process has brought enormous benefits to quality, availability, realism, and personalization. On the other hand, it has raised issues at multiple levels \cite{jiang2023ai}. In particular, concerns about the misuse of generative AI have increased at the same fast pace \cite{marchal2024generative}. As a consequence, efforts for the detection of AI-generated images (often referred to as fake images) and the identification of their source have increased accordingly \cite{tariang2024synthetic}.

Among these efforts, the goal of source attribution consists in identifying which generator produced a synthetic image \cite{yu2019attributing, marra2019gans}. This problem can be tackled together with detection or as a standalone task. Our work focuses on the latter setting: we assume to already know that a given image is synthetic and wish to determine the model that was used to generate it.

Synthetic Image Attribution (SIA) is made even more challenging by the sheer number of sources available online and by the large number of new ones published every month \cite{zhang2024texttoimagesynthesisdecadesurvey}. It is, then, unfeasible to train a new SIA model every time a new generator appears. In addition, acquiring large amounts of data from commercial generators to fully train a SIA model on them is very expensive. For these reasons, few-shot attribution, where only a small number of examples per source is used for training, is attracting increasing interest \cite{sheng2022fewshot}. This task can be addressed using a classifier trained on a handful of examples per class \cite{defake}, or with a modular approach that attributes images by clustering, distance, or similarity \cite{bindini2024tiny}.

In this paper, we introduce a training-free method for SIA based on resynthesis, which works in a one-shot setting. The synthetic image is assumed to be from a generator within a set of known sources. A textual description of the image is obtained with a captioning model and used to produce a \textit{resynthesis} of the to-be-attributed image with each of the candidate sources. The \textit{resynthesis} which is most similar to the \textit{original} is assumed to be the one generated by the same source. The input image is then attributed accordingly \cite{albright2019source}. To judge image similarity, given the impossibility of using pixel-wise distances \cite{marra2019neural}, we resort to a combination of high-level semantic features and low-level signatures \cite{marra2019neural}, extracted with a pretrained CLIP (Contrastive Language Image Pretraining) model \cite{clip, cioni2024clipfeaturesneeduniversal}. The distance is measured in the feature space using a simple distance function.

\begin{figure*}[t]
\centering
\includegraphics[width=\linewidth]{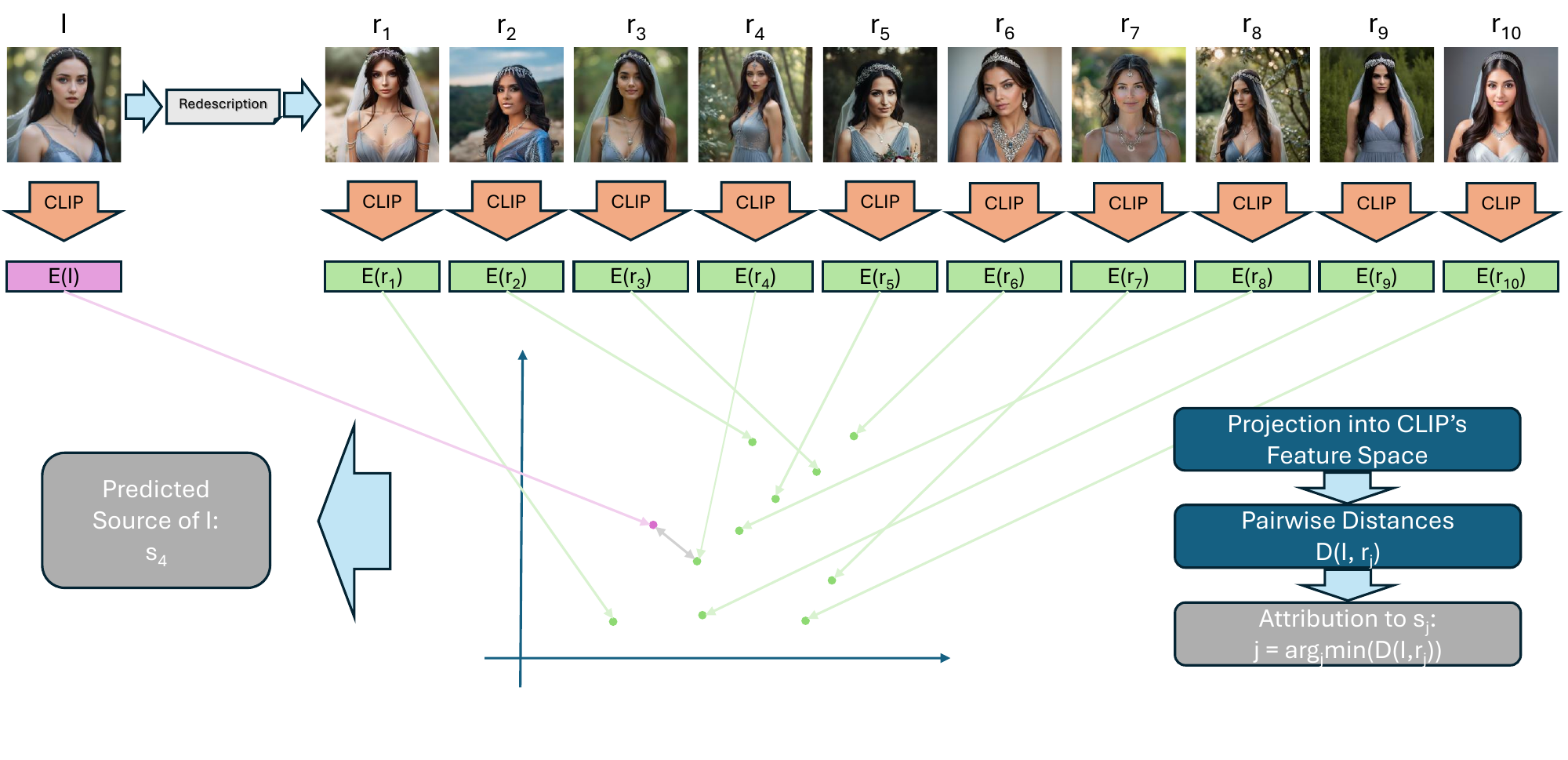}
\caption{Workflow of the proposed source attribution method based on resynthesis. The \textit{original} image $I$ in this example was generated by Freepik. From left to right, the \textit{resyntheses} $r_{j}$ were produced by: Bing, Firefly, Flux-dev, Freepik, Imagen3, Leonardo AI, Midjourney, Nightcafe, Stable Diffusion 3, Starry AI. The features of $I$ and all the $r_{j}$ are extracted with CLIP. The distances between the projections $E(I)$ and each $E(r_j)$ in CLIP's feature space are measured, and $I$ is attributed to the source $s_{j*}$ with the minimum distance $d(E(I), E(r_{j*}))$. In this example, the image is correctly attributed to $s_4$, which corresponds to Freepik.}
\label{fig:method}
\end{figure*}

Prior work has introduced various benchmarks for SIA \cite{goebel2022synthetic, frank2020leveraging, gragnaniello2020detection}. However, most datasets are limited to open-source generators or to a small number of sources, limiting their potential to represent real-world settings where many generative methods, including commercial ones, are available to the average user. Moreover, existing datasets often do not include \textit{resyntheses}, making it difficult to train and test distance-based or similarity-based methods \cite{carlini2021difficulty}. Hence, as a second contribution, we propose a new dataset incorporating the resynthesis mechanism. The dataset focuses on head-and-shoulder photo-portrait style images produced by text-to-image generators. It incorporates two key features: i) images generated by a set of 14 sources, including 7 commercial generators; ii) secondary descriptions used to generate \textit{resyntheses} of each image, allowing the use of resynthesis-based methods. We used the new dataset to test the performance of our source attribution method and to compare it with several baselines, establishing benchmarks for future research. In summary, the main contributions of our work are:
\begin{itemize}
\item a training-free, one-shot source image attribution method based on image resynthesis;
\item a dataset for few-shot SIA, incorporating a resynthesis mechanism based on text image description and generation of \textit{resyntheses};
\item a set of experiments including comparisons with state-of-the-art few-shot source attribution methods, to assess the performance of our method and validate its superiority in a data scarcity regime, also setting baselines for future research on the new dataset.
\end{itemize}

\begin{figure*}[!ht]
\centering
\includegraphics[width=\linewidth]{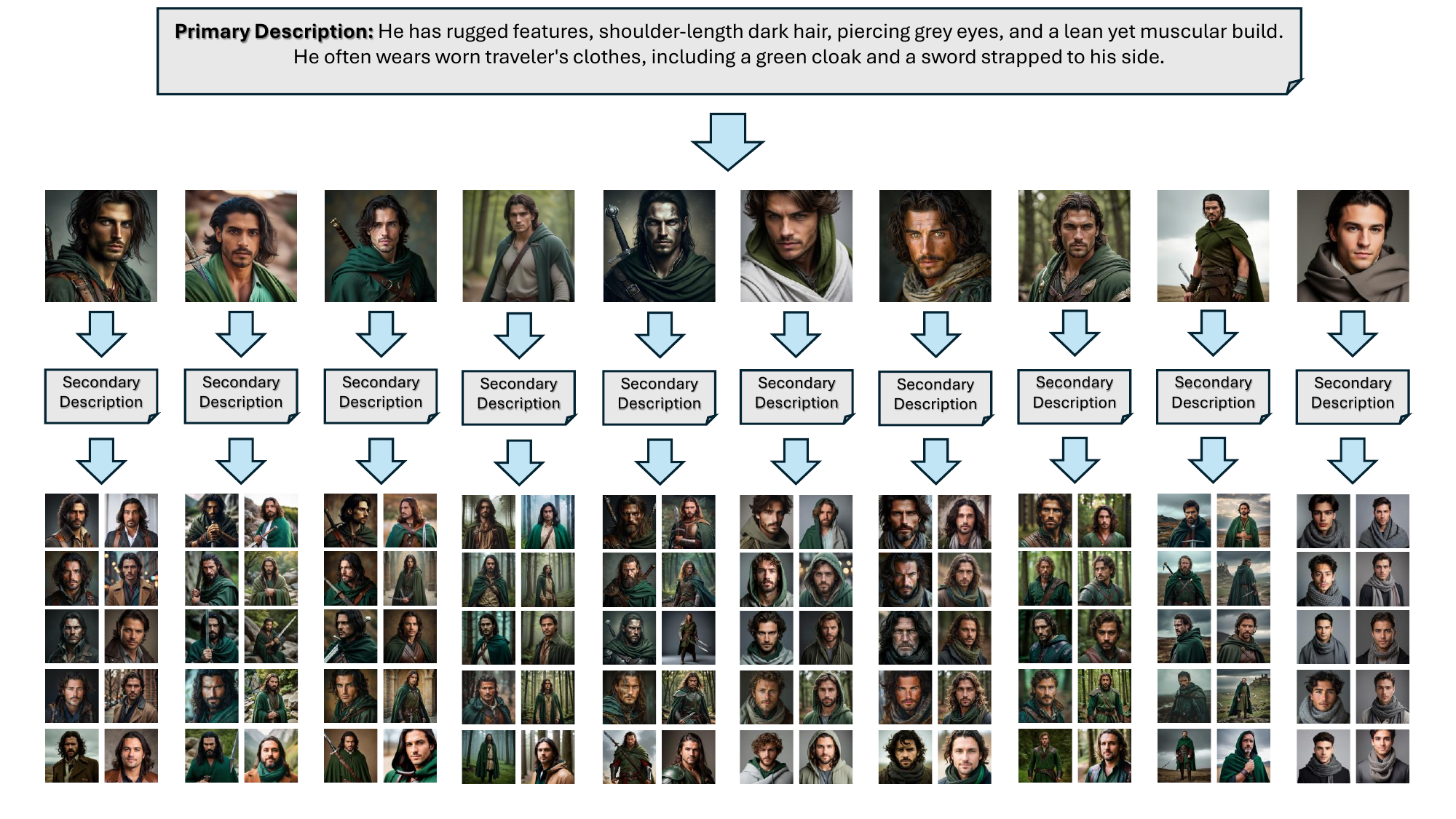}
\caption{Dataset building procedure. This example corresponds to \textit{character} \#$42$: \textit{Aragorn (The Lord of The Rings)}. Starting from the top and going down: The primary description is used as a prompt for the 10 core generative models introduced in Section \ref{subsec:generators}, which produce the images in the first row. The models are in alphabetical order, from left to right. Each image is then passed to Chat-GPT to obtain a secondary description. The latter is used to generate the \textit{resyntheses} in the bottom double-column groups, using all the 10 core generators again. In each group of \textit{resyntheses}, from left to right and from top to bottom, the generators are again in alphabetical order.}
\label{fig:building_schema}
\end{figure*}

\section{Method}
\label{sec:method}

Resynthesis is used for image classification in various settings \cite{dong2025prsn}. In a real-world case, the prompt used to generate a synthetic image is usually unavailable. Given an image to attribute, our approach relies on a secondary description of the \textit{original} image to build \textit{resyntheses} (resynthesized versions of the image) with all the candidate sources. The description is obtained with Chat-GPT \cite{chat-gpt} and has a maximum length of 200 characters. In such panel of \textit{resyntheses}, the \textit{resynthesis} which mostly resembles the \textit{original} image should be the one generated by the same model. As a consequence, attribution can be obtained by distance calculation: the \textit{original} image is attributed to the generator that produced the closest \textit{resynthesis}. Formally, given a set of $n$ candidate text-to-image sources $S = \{s_1, \dots s_n\}$, and a distance function $d$, an image $I$ is attributed to the source $s_j$ by extracting a prompt $p(I)$ describing the content of $I$, then obtaining a set of \textit{resyntheses} $\{r_j = s_j(p(I)) \ \forall s_j \in S\}$, and finally calculating the distances $\{d_j = d(r_j, I) \ \forall s_j \in S\}$. The image $I$ is attributed to the source $s_{j^*}$ achieving the minimum distance. A pre-trained CLIP Large model (input size $336\times336$) \cite{clip_large_hf} is used as a feature extractor and the distance is measured in the corresponding feature space. Defining the feature extraction as an operator $E()$ and indicating by $d()$ the distance function, we have:

\begin{equation}
    \label{eq:extract_attribute}
    s_{j^*} : j^* = \arg\min_j (d( E(s_j(p(I))), E(I) )).
\end{equation}

We tested several distances $d$, including Euclidean, Manhattan, Mahalanobis, and Cosine distances. The CLIP model is frozen, we do not fine-tune it. Hence, our model is training-free, and can be considered a one-shot method. The only operations required to attribute an image, in fact, are the extraction of the secondary description with a generic image-to-text language model (chat-GPT in our case \cite{chat-gpt}), and the generation of one \textit{resynthesis} for each candidate source. A summary of our methodology is shown in Figure \ref{fig:method}.

\section{Dataset}
\label{sec:dataset}
To assess the performance of the proposed SIA method and compare it with few-shot baselines, we created a new dataset\footnote{The dataset will be made publicly available upon publication of the paper.}. The dataset has innovative features including: i) images generated by 14 generators (10 core sources, and 4 for the dataset extension), ii) 7 commercial generators, iii) integration of the resynthesis mechanism with the inclusion of \textit{original} images and their corresponding \textit{resyntheses}. The dataset contains 12,000 head-and-shoulder images of novel characters, produced with 14 text-to-image generators described in Section \ref{subsec:generators}.

\subsection{Generators}
\label{subsec:generators}
The sources used to generate the dataset comprise 7 open-source models (4 of which were used for the dataset extension) and 7 commercial generators. They are listed in the following, with the * symbol indicating commercial models, and the \textsuperscript{e} symbol the models used in the dataset extension:
\begin{itemize} 
    \item Bing (Dall-E 3) * \cite{dalle3}
    \item Firefly * \cite{adobe_firefly}
    \item Flux.1.dev \cite{flux_dev}
    \item Freepik \cite{freepik}
    \item Imagen3 * \cite{imagen3}
    \item Leonardo AI * \cite{leonardoai}
    \item Midjourney * \cite{midjourney}
    \item Nightcafe * \cite{nightcafe}
    \item Stable Diffusion 3 \cite{sd3}
    \item Starry AI * \cite{starryai}
    \item AuraFlow \textsuperscript{e} \cite{auraflow}
    \item Pixart \textsuperscript{e} \cite{pixart}
    \item Playground v2.5 \textsuperscript{e} \cite{playground}
    \item Tencent Hunyuan \textsuperscript{e} \cite{tencent_hunyuan}
\end{itemize}

\begin{figure}[t]
\centering
\includegraphics[width=1.0\linewidth]{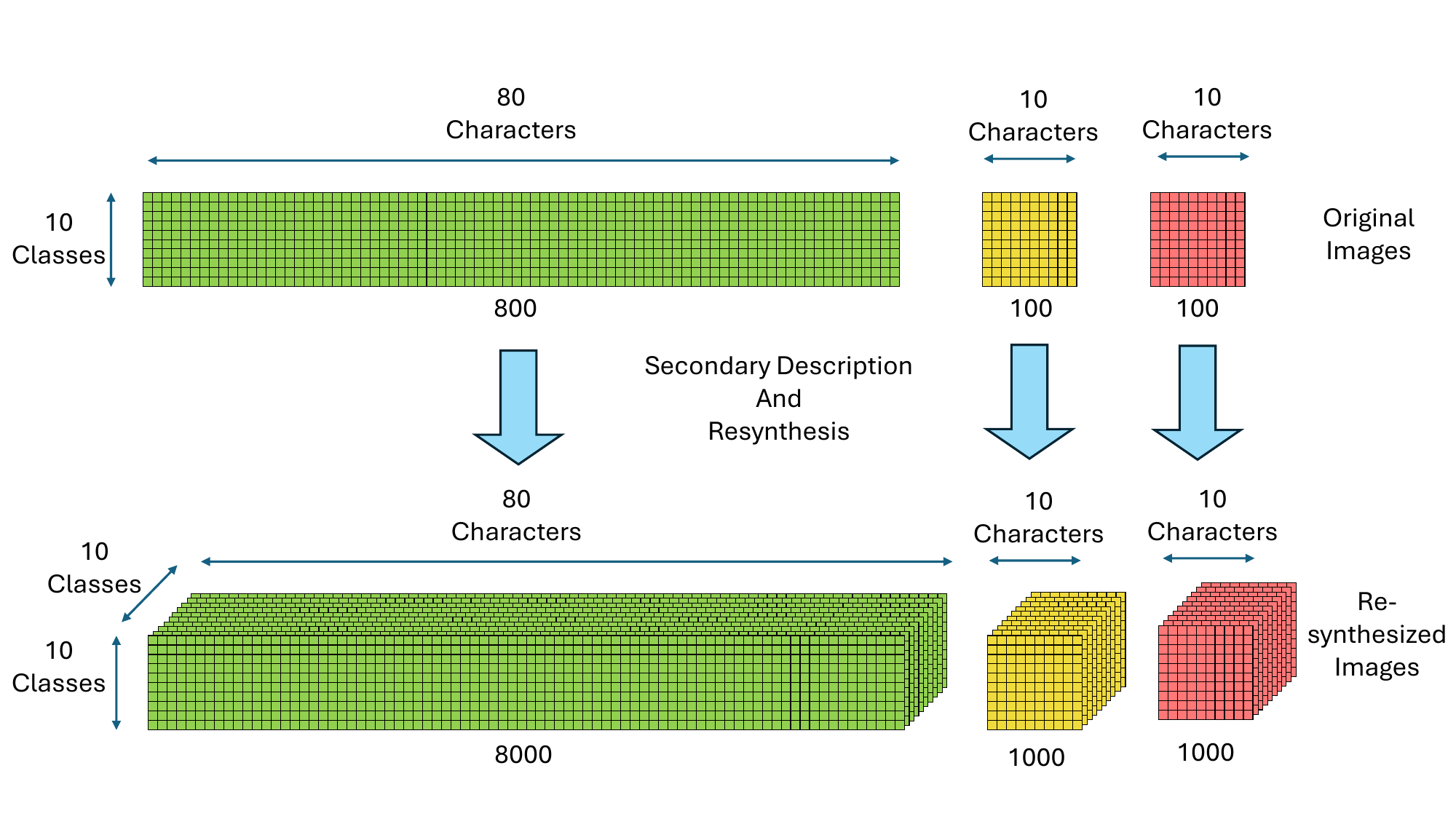}
\caption{Dataset composition and split. Each of the $100$ \textit{characters} is generated with all the sources, obtaining $1,000$ \textit{original} images. Their secondary descriptions are extracted with ChatGPT and used to generate $10$ \textit{resyntheses} each, one with each source. These are the components of the main dataset. The dataset is split along the \textit{character} index, meaning that all the $10$ \textit{original} images of each character and their $100$ \textit{resyntheses} belong to the same set. The dataset extension (Ex) is then obtained by taking the test set and adding the other $4$ sources. $40$ new \textit{original} images are produced in this way, along with their $40\times14=560$ \textit{resyntheses} and $400$ \textit{resyntheses} of the existing \textit{original} images with the $4$ new sources.}
\label{fig:diagram}
\end{figure}

\subsection{Dataset Construction}
\label{subsec:building}
We used $100$ descriptions of novel characters summarized with ChatGPT \cite{chat-gpt} to get prompts of uniform length (max $200$ character). We built an index on the \textit{character} axis and used it to split the dataset. This prevents images originating from the same prompt ending up in different sets and guarantees that the classes are always balanced. For each prompt, we generated $10$ \textit{original} images, one with each of the core sources described in Section \ref{subsec:generators}, obtaining  $1,000$ \textit{original} images. These are linked into groups of $10$ images produced from the same description by the \textit{character} index. To generate the \textit{resyntheses}, a \textit{secondary} description for each \textit{original} image was produced by Chat-GPT \cite{chat-gpt} using the command: "Provide a detailed description of the character represented in this image in less than 200 characters". \textit{Original} images of the same character from different sources have different \textit{secondary} descriptions, based on the semantical elements introduced by the models and on low-level signatures that can influence language model style. For each secondary description, we generated $10$ \textit{resyntheses} (one with each core model). The $10$ \textit{resyntheses} of each \textit{original} image come from the same secondary description and are linked. They are also linked to the other $9$ \textit{original} images from the same \textit{primary} description and to all their \textit{resyntheses}, constituting groups of $100$ \textit{resyntheses} originating from the same prompt. We split the dataset in training, validation and test sets of $80$, $10$, and $10$ characters. Consequently, the training set is composed of $800$ \textit{original} images (OrTr) and their $8,000$ \textit{resyntheses} (ResTr), while the validation and test sets are composed each of $100$ \textit{original} images (OrVa, OrTe) and their $1,000$ \textit{resyntheses} (ResVa, ResTe). The procedure is shown in Figure \ref{fig:building_schema}. For few-shot experiments on more than 10 classes, we built an extension with $4$ additional sources, using only the $10$ test set characters. We have $40$ additional \textit{original} images, their $560$ \textit{resyntheses} generated with all the $14$ models, and $400$ \textit{resyntheses} of the $100$ core \textit{original} images of the test set obtained with the new sources. These are added to the test set, obtaining the extended dataset, which totals $140$ \textit{originals} (OrEx) and their $1960$ \textit{resyntheses} (ResEx). A scheme is shown in Figure \ref{fig:diagram}.

\section{Experiments}
\label{sec:experiments}

\subsection{Baselines}
\label{subsec:baselines}

\subsubsection{CLIP Feature Classifiers}
\label{subsubsec:clip_plus}
In addition to resynthesis, we use CLIP-Large (input size $336\times336$) to get image features which are fed into a small classifier: either a Multi-Layer Perceptron (CLIP+MLP) or a Support Vector Machine (CLIP+SVM). The CLIP parameters are frozen and not fine-tuned. The hyperparameters of both heads were optimized on the validation set. For CLIP+MLP, the architecture has two hidden layers with $512$ and $32$ units, ReLU activation and a softmax output layer. In few-shot scenarios, the architecture is reduced to a single hidden layer of $512$ units. The MLP was trained using Adam \cite{Adam}, with initial learning rate $10^{-3}$, and $l_{2_\alpha}$ set to $10^{-4}$, for a maximum of $1,000$ epochs. Early stopping was applied on the validation loss with a patience of 10 epochs. CLIP+SVM employed a Radial Basis Function (RBF) kernel with regularization $C=1.0$, tolerance $10^{-3}$, kernel degree $3$, and no iteration limit.

\subsubsection{De-Fake}
\label{subsubsec:defake}
De-Fake \cite{defake} leverages subtle artifacts left by generators for SIA. The architecture comprises a detector and an attributor. We use only the attributor module, as all images in our dataset are synthetic. All experiments were conducted using the default parameters reported in \cite{defake}.

\subsubsection{CLIP-LoRA}
\label{subsubsec:cliplora}
CLIP-LoRA \cite{zanella2024low} applies a low-rank adaptation to CLIP to fine-tune it on few-shot tasks. We apply the LoRA adapter to CLIP Large. The classification head is a MLP with two hidden layers of $768$ and $1024$ units respectively, both with ReLU activation. The model is fine-tuned for $12$ epochs ($10$ in few-shot experiments) using Adam \cite{Adam}, initial learning rate $10^{-3}$, and batch size $16$.

\begin{figure*}[ht]
\centering
    \subfloat[5 Classes\label{1a}]{%
       \includegraphics[width=0.3\linewidth]{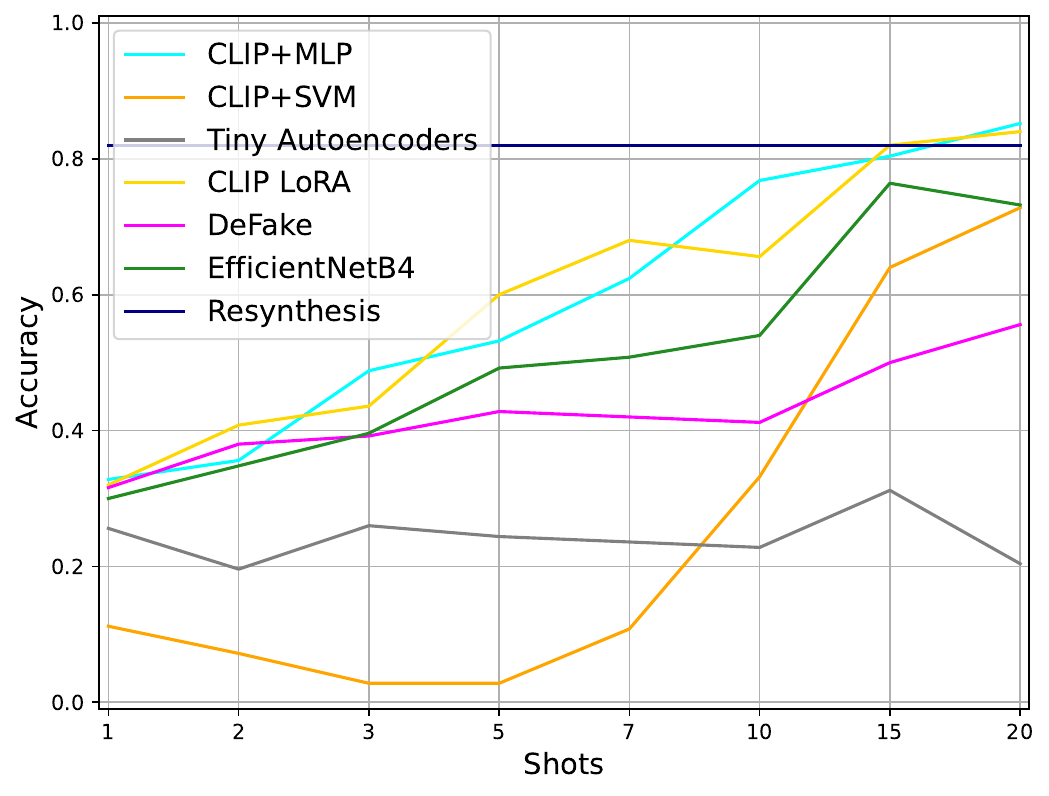}}
    \hfill
    \subfloat[8 Classes\label{1b}]{%
       \includegraphics[width=0.3\linewidth]{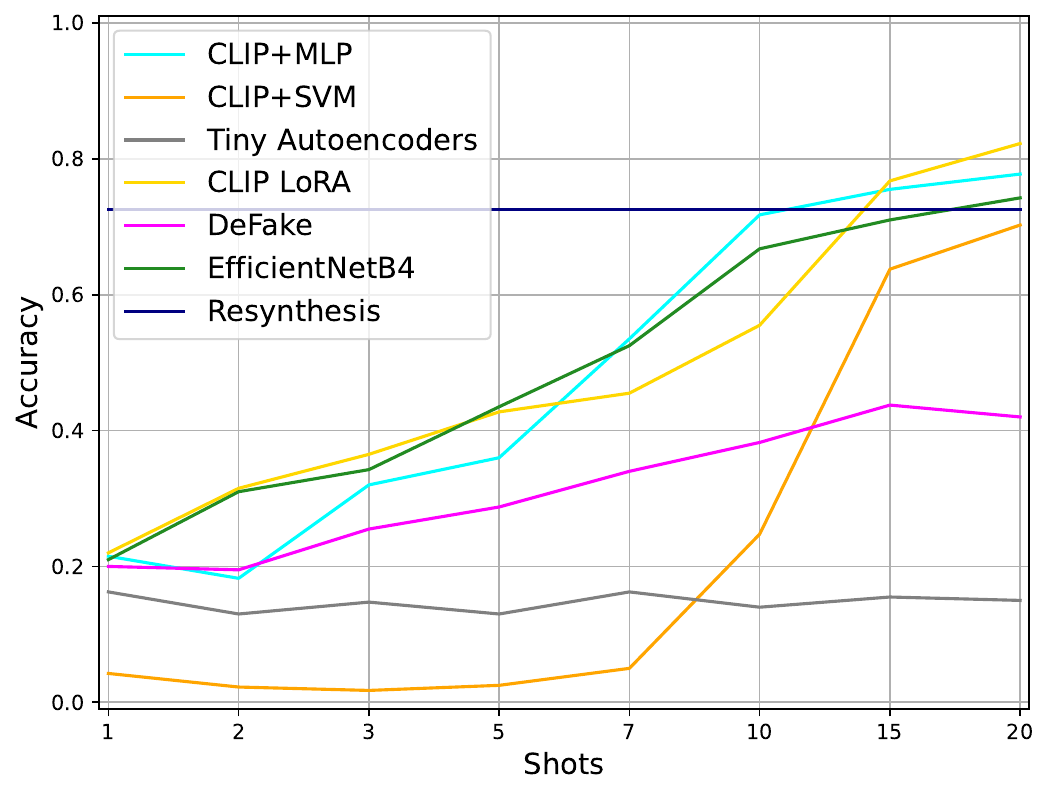}}
    \hfill
    \subfloat[10 Classes\label{1c}]{%
       \includegraphics[width=0.3\linewidth]{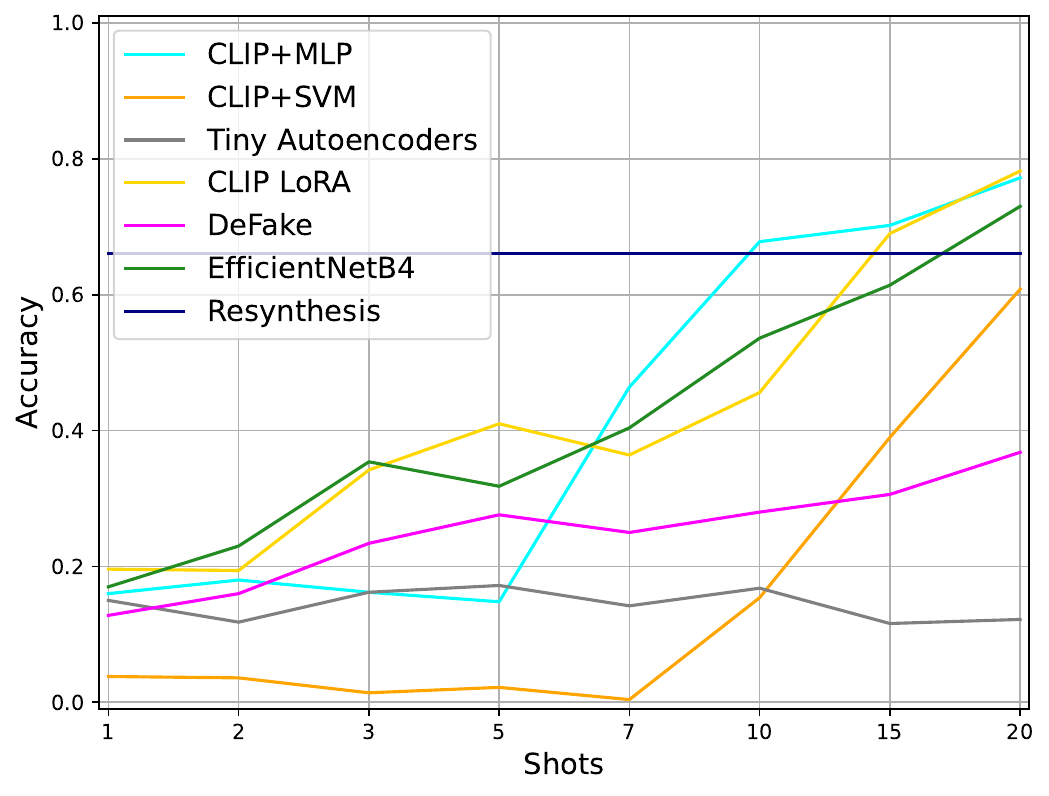}}
    \\
    \hfill
    \subfloat[12 Classes\label{1d}]{%
       \includegraphics[width=0.3\linewidth]{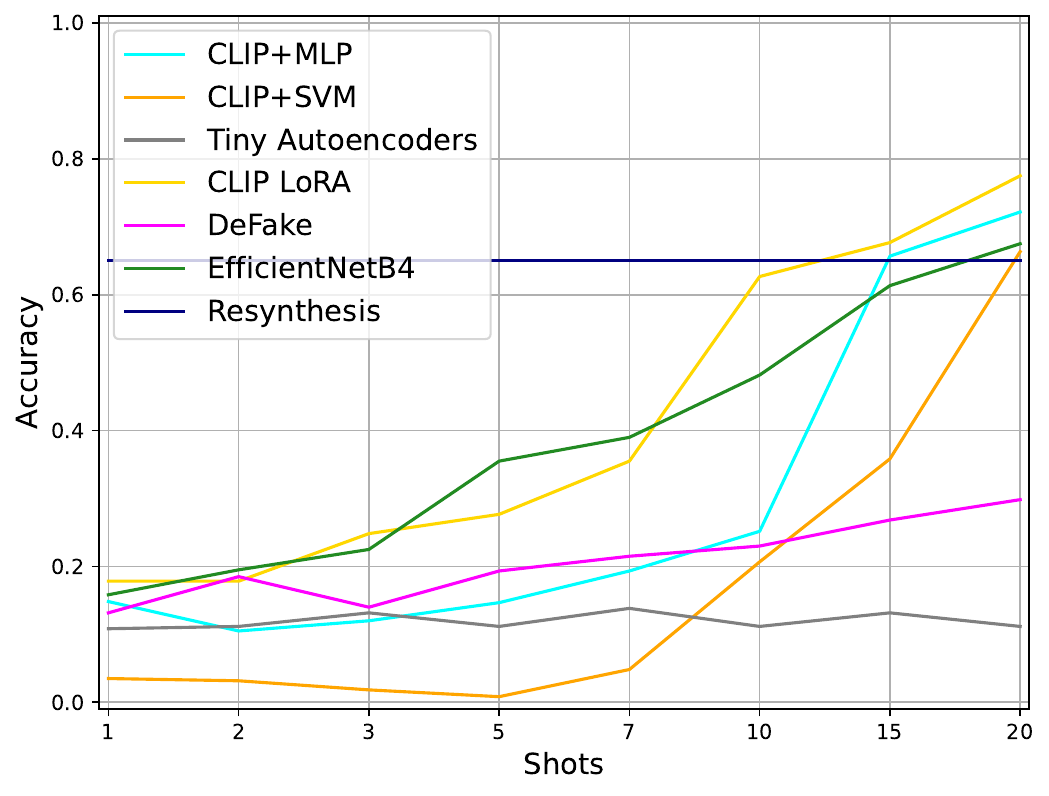}}
    \hfill
    \subfloat[14 Classes\label{1e}]{%
       \includegraphics[width=0.3\linewidth]{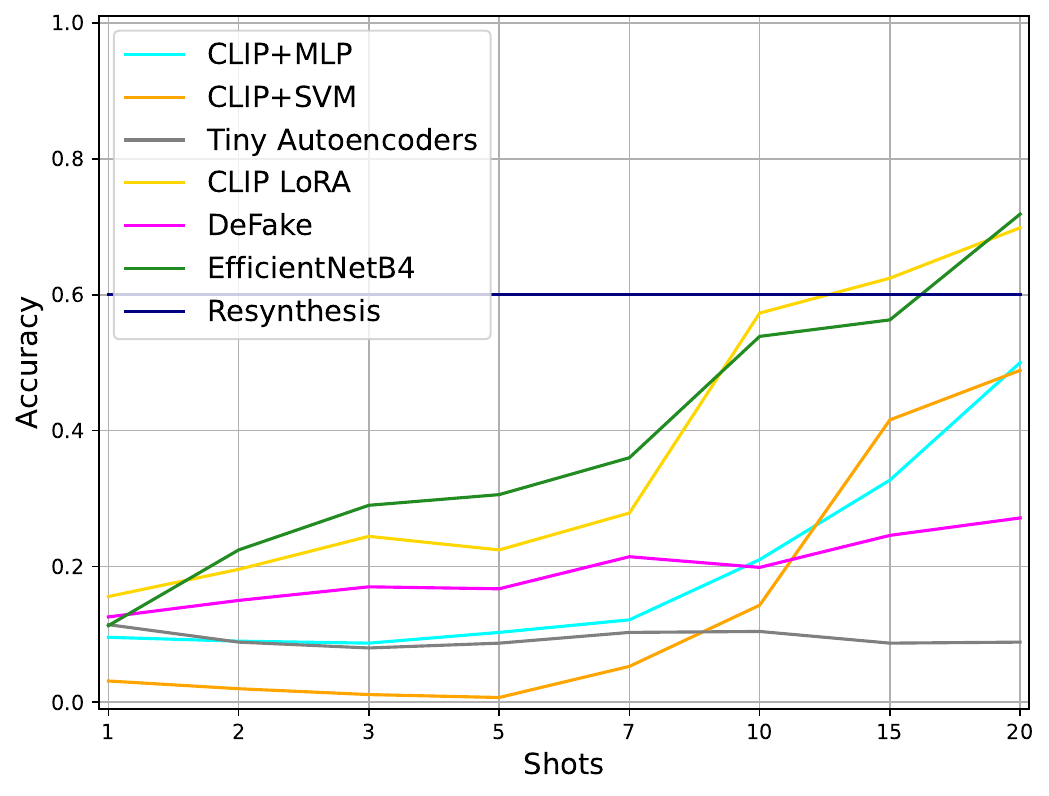}}
    \hfill
\caption{Accuracy in a few-shot scenario (T1) with different numbers of classes. For the experiments with 12 and 14 classes, the test-only sources were used. The Accuracy is plotted with respect to the number of shots available for training. Our Resynthesis method is the navy-blue line. The performance of our training-free method does not depend on the number of shots, and is therefore constant. The intersection between the line representing our method's accuracy and a baseline's accuracy line corresponds to the number of shot needed by that baseline to outperform our training-free (one-shot equivalent) method.}
\label{fig:few_shot}
\end{figure*}

\begin{table*}[htbp]
\caption{Accuracy on task T1 - plain source attribution, and T2 - robust source attribution. Each column corresponds to a different post-processing operator applied to test set images. The best result of each column is highlighted in bold.}
\label{tab:accuracy}
\tiny
\begin{center}
\begin{tabular}{|c|c|cccccccccc|} 
\hline
Task & T1 & \multicolumn{10}{|c|}{T2} \\
Method & Plain & Blur & Brightness & Contrast & Crop & Greyscale & JPEG & Resize & Rotation & Social & WEBP \\
\hline
CLIP+MLP & $0.95$ & $0.78$ & $0.88$ & $0.65$ & $0.81$ & $0.72$ & $0.68$ & $0.95$ & $0.82$ & $\mathbf{0.81}$ & $0.68$\\
CLIP+SVM & $\mathbf{0.97}$ & $\mathbf{0.86}$ & $\mathbf{0.94}$ & $\mathbf{0.74}$ & $\mathbf{0.82}$ & $\mathbf{0.74}$ & $0.70$ & $\mathbf{0.97}$ & $\mathbf{0.83}$ & $0.80$ & $0.73$ \\
Tiny Autoencoders \cite{bindini2024tiny} & $0.16$ & $0.12$ & $0.15$ & $0.12$ & $0.03$ & $0.14$ & $0.16$ & $0.16$ & $0.15$ & $0.16$ & $0.16$\\
CLIP-LoRA \cite{zanella2024low} & $0.78$ & $0.77$ & $0.62$ & $0.63$ & $0.50$ & $0.62$ & $0.70$ & $0.79$ & $0.77$ & $0.75$ & $\mathbf{0.74}$\\
De-Fake \cite{defake} & $0.72$ & $0.65$ & $0.67$ & $0.68$ & $0.72$ & $0.64$ & $\mathbf{0.72}$ & $0.72$ & $0.70$ & $0.71$ & $0.72$\\
EfficientNetB4 \cite{tan2019efficientnet} & $0.61$ & $0.63$ & $0.57$ & $0.60$ & $0.37$ & $0.64$ & $0.59$ & $0.62$ & $0.61$ & $0.60$ & $0.60$\\
Resynthesis (ours) & $0.66$ & $0.62$ & $0.62$ & $0.52$ & $0.50$ & $0.54$ & $0.53$ & $0.67$ & $0.62$ & $0.58$ & $0.47$\\
\hline
\end{tabular}
\end{center}
\end{table*}

\subsubsection{EfficientNetB4}
\label{subsubsec:efficientnet}
Convolutional Neural Networks (CNNs) are usually the best option for image classification. The CNN which usually shows the best performance in SIA \cite{mandelli2024synthetic} is EfficientNetB4 \cite{tan2019efficientnet}. This model is not easy to fine-tune without a large number of samples per class. Hence, our dataset is challenging for it. We fine-tuned EfficientNetB4 for $12$ epochs ($10$ in few-shot experiments), using Adam \cite{Adam}, initial learning rate $10^{-4}$, and batch size $16$.

\subsubsection{Tiny Autoencoders} 
\label{subsubsec:tae}
Tiny Autoencoders \cite{bindini2024tiny} is a modular SIA method. For each suspect source, a small CNN autoencoder is trained on a handful of images generated by that source. We used the standard parameters introduced in \cite{bindini2024tiny}.

\subsection{Tasks and settings}
\label{subsec:tasks}
The \textit{resyntheses} are used for attributing the \textit{originals} based on resynthesis. The methods are evaluated using the Accuracy metric. Tests were carried out on the following tasks.

\paragraph{T1 - Few-Shot SIA} only a handful of examples are available to train the methods. The experiments are carried out with different numbers of shots (1,2,3,5,7,10,15,20) and different numbers of classes (5,8,10,12,14). All the tests were carried out on the dataset extension. We averaged the results on 5 repetitions of the experiment, each with a different seed for weight initialization and with different training examples (the same for all the models). 

\paragraph{T2 - Plain SIA} consists in evaluating SIA performance on OrTe dataset, exploiting OrTr, ResTr, OrVa, and ResVa to train the trainable models.

\paragraph{T3 - Robust SIA} similar to T2, the only difference being that the test images are post-processed with a single operator at a time. None of the methods use data augmentation with post-processing operations during fine-tuning. Ten different tests are carried out, using a different post-processing operator each time: Blur; Brightness (1.2 to 2.4 increase), Contrast (1.2 to 2.4 increase); Crop (central, 0.5 to 0.9 the original area); Greyscale; JPEG ($QF\in[50,99]$); Resize (0.4 to 2.0 the original size); Rotation (-5° to +5°, aspect preserving); Social (simulating an upload on Instagram); WEBP ($QF\in[50,99]$). Every operator except Blur and Greyscale has a parameter adjusting the transformation strength: its value is randomized for every image, but the same value is used for all the methods.

For task T1 (few-shot), the training samples are limited to the number of shots. Let $k$ be the number of shots, the training ssamples are randomly drawn among the resyntheses of $k$ characters (the same number for each source), while the other 100-$k$ characters are used for testing. Our resynthesis approach does not use the training samples, attributing the \textit{original} images by measuring the distance from their \textit{resyntheses}, in a one-shot fashion. In the following, we report only the results obtained with the Euclidean distance, as other distances yielded worse performance. While task T1 is the main focus of our experimentation, T2 and T3 are presented for a reference performance in a regular training scenario, where few-shot is not needed. In the latter, the baselines can be trained on OrTr, ResTr, or both. The methods requiring training or fine-tuning were trained once and then tested on both T2 and T3. For each method, we trained three different versions: one using OrTr, one using ResTr, and one using both. The best version of each model was selected based on validation accuracy. For De-Fake, the best version was the one trained on OrTr, while for all the other methods the best version was the one trained on ResTr. The results refer to these versions.

\subsection{Results and Discussion}
\label{subsec:results}
The results of task T1 are reported in Figure \ref{fig:few_shot}. Our method outperforms the baselines when few examples are available, on all the numbers of classes. Resynthesis is the best method when 10 or less shots per class are available. Since our method is computationally equivalent to a one-shot method, it is always the best choice for 10 or less shots. This is shown by the navy-blue line representing the performance of resynthesis: line crossing always occurs at 10 or 15 shots. When more shots are available, the best methodologies are CLIP LoRA and CLIP+MLP (the latter shows bad performance on 14 classes).
Table \ref{tab:accuracy} displays Accuracy on tasks T2 and T3. The best method is CLIP+SVM, with CLIP+MLP following close. Their advantage comes from the fact that CLIP weights are not fine-tuned, with all the training being dedicated to the SVM/MLP. This, combined with the high informative potential of CLIP features on both low-level and high-level features, allows these two simple models to obtain the best results. The SVM has a well-known advantage in terms of robustness, which makes it outperform the other methods also on the various operators of task T3. The only exceptions are JPEG (De-Fake), WEBP (CLIP-LoRA), and Social (CLIP+MLP). CLIP-LoRA and De-Fake, in particular, show high-robustness. The fact that also these two methods are based on CLIP reinforces the idea that CLIP is capable of embedding high-quality information at all levels. Unsurprisingly, in these tasks, resynthesis is not capable of competing with models which are trained or fine-tuned, even if its performance is not bad.

\section{Conclusions}
\label{sec:conclusion}
We presented a new training-free method for SIA based on resynthesis, and a dataset for few-shot SIA that contains secondary descriptions and image \textit{resyntheses}. Results show that our method has very good performance with respect to state-of-the-art baselines, outperforming them consistently when 10 or less shots are available for training. Since our algorithm needs to generate just one \textit{resynthesis} per class, it is equivalent to a one-shot method, and hence constitutes the best attribution method when the number of shots is between 1 and 10. This was demonstrated with different numbers of sources, ranging between 5 and 14. We also introduced a new dataset which proved to be a challenging benchmark for few-shot SIA. Future work will investigate more sophisticated distance functions for the resynthesis method and analyze the impact of the secondary description on overall performance. We will also assess whether ChatGPT is the most suitable choice for generating secondary descriptions when the original prompt is produced by a different method, and explore alternative models for this task. In addition, new image categories will be incorporated into the dataset, and resynthesis performance will be evaluated on these expanded categories.

\section*{Acknowledgment}
This work was partially supported by project SERICS (PE00000014) under the MUR National Recovery and Resilience Plan funded by the European Union - NextGenerationEU.

\bibliographystyle{ieeetr}
\bibliography{bibliography}

\end{document}